\documentclass[12pt]{article}
\usepackage{amsmath}
\usepackage{graphicx}
\usepackage{natbib}
\usepackage{url} % not crucial - just used below for the URL 

\newcommand{\blind}{0}

\usepackage{amsmath,amsfonts,amssymb,amsthm,mathtools}

\usepackage[colorlinks=true,citecolor=blue]{hyperref}
\usepackage{numprint}
\usepackage[a4paper, total={6in, 8in}]{geometry}
\usepackage{authblk}
\usepackage{comment}
\usepackage{xcolor}
\usepackage{acronym}

\usepackage{bbm}
\theoremstyle{plain}
\newtheorem{assumption}{Assumption}
\newtheorem{theorem}{Theorem}
\newtheorem{remark}{Remark}

\newtheorem{prop}{Proposition}
\newtheorem*{assumption*}{Assumptions}
\newcommand\norm[1]{\left\lVert#1\right\rVert}

\newcommand{\indep}{\;\, \rule[0em]{.03em}{.67em} \hspace{-.25em}\rule[0em]{.65em}{.03em}\hspace{-.25em}\rule[0em]{.03em}{.67em}\;\,}

\DeclareMathOperator*{\argmin}{arg\,min}

\DeclarePairedDelimiter\floor{\lfloor}{\rfloor}

\usepackage{enumerate}

\newacro{ReLU}{Rectified Linear Unit}
\newacro{CNN}{Convolutional Neural Networks}
\newacro{MLP}{Multilayer Perceptron}
\newacro{AIPW}{Augmented Inverse Probability Weighting}
\newacro{FNN}{Feed-forward Neural Networks}
\newacro{DR}{Double Robust}
\newacro{ACE}{Average Causal Effect}
\newacro{ACET}{Average Causal Effect among the Treated}
\newacro{OR}{Outcome Regression}
\newacro{TMLE}{Targeted Maximum Likelihood Estimation}
\newacro{DGP}{Data Generating Process}

\begin{document}

\def\spacingset#1{\renewcommand{\baselinestretch}%
{#1}\small\normalsize} \spacingset{1}

%%%%%%%%%%%%%%%%%%%%%%%%%%%%%%%%%%%%%%%%%%%%%%%%%%%%%%%%%%%%%%%%%%%%%%%%%%%%%%

\if0\blind
{
  \title{\bf Convolutional neural networks for valid and
efficient causal inference}
  \author{Mohammad Ghasempour\thanks{We are grateful to Xijia Liu and Jenny Häggström for helpful comments that have improved the paper. We acknowledge funding from the Swedish Research Council and the Marianne and Marcus Wallenberg Foundation. The Umeå SIMSAM Lab data infrastructure used in this study was developed with support from the Swedish Research Council, the Riksbanken Jubileumsfond and by strategic funds from Umeå University.\\
  Correspondence: mohammad.ghasemour@umu.se}, Niloofar Moosavi, Xavier de Luna\hspace{.2cm}\\
    Department of Statistics, USBE, Umeå University, Umeå, Sweden \\
    }
  
  \maketitle
} \fi

\if1\blind
{
  \bigskip
  \bigskip
  \bigskip
  \begin{center}
    {\LARGE\bf Title}
\end{center}
  \medskip
} \fi

\bigskip
\begin{abstract}
Convolutional neural networks (CNN) have been successful in machine learning applications. Their success relies on their ability to consider space invariant local features. We consider the use of CNN to fit nuisance models in semiparametric estimation of the average causal effect of a treatment. In this setting, nuisance models are functions of pre-treatment covariates that need to be controlled for. In an application where we want to estimate the effect of early retirement on a health outcome, we propose to use CNN to control for time-structured covariates. Thus, CNN is used when fitting nuisance models explaining the treatment and the outcome. These fits are then combined into an augmented inverse probability weighting estimator yielding efficient and uniformly valid inference. Theoretically, we contribute by providing rates of convergence for CNN equipped with the rectified linear unit activation function and compare it to an existing result for feedforward neural networks. We also show when those rates guarantee uniformly valid inference. A Monte Carlo study is provided where the performance of the proposed estimator is evaluated and compared with other strategies. Finally, we give results on a study of the effect of early retirement on hospitalization using data covering the whole Swedish population.
\end{abstract}

\noindent%
%{\it Keywords:}  Average causal effect; augmented inverse probability weighting; early retirement; post-machine learning inference.
\vfill

\newpage
\spacingset{1} % DON'T change the spacing!
\section{Introduction}
\label{sec:intro}

\ac{CNN} have been found successful in discovering location-invariant patterns in speech, images, and time series data \citep{lecun1995convolutional}. In particular, they have been shown to have a universal approximation property and be more efficient in terms of number of hidden layers than fully-connected multi-layer networks in high-dimensional situations \citep{zhou2020universality}. In this paper we show how \ac{CNN} can be useful in controlling confounding information when using rich observational databases in order to perform semiparametric inference on a low dimensional causal parameter: we focus on average causal effects of a binary treatment on an outcome of interest, although our results are relevant for the semiparametric estimation of other low dimensional parameters of interest \citep[see e.g.][]{chernozhukov2018double}.

\ac{AIPW} estimators \citep[also called \ac{DR} estimators,][]{robins1994estimation}  attain the semiparametric efficiency bound and yield uniformly valid inference as long as the nuisance functions of the confounding covariates are fitted consistenlty with fast enough convergence rates; e.g. all nuisance functions are estimated with order $n^{-1/4}$ \citep{belloni2014inference,Farrell_2015,kennedy2016semiparametric,moosavi2021costs}. In this paper, we contribute to this theory by showing that \ac{CNN} fits of nuisance functions achieve the $n^{-1/4}$ convergence rate required to obtain uniformly valid inference on causal parameters. To show this we use a result obtained by \citet{farrell2018deep} for general \ac{ReLU}-based \ac{FNN}. They show that, for large samples, the estimation error rate of \ac{FNN} are bounded by the following term, with a probability increasing exponentially with $\gamma$: 
    \begin{flalign}\label{convergencebound}
      & \sqrt{ \log n/n} \times \text{complexity penalty} + \sqrt{(\log \log n + \gamma)/n} + \text{approximation rate}.
    \end{flalign}
We deduce the above approximation rate for \ac{CNN} architectures inspired by earlier work by \citet{zhou2020universality}. However, in contrast to the latter paper, we consider a larger number of free parameters by considering multi-channel convolutional neural network, so as to achieve a trade-off between complexity penalty and approximation rate in \eqref{convergencebound}, and thereby obtain the convergence rate $n^{-1/4}$ for the \ac{CNN} fit of the nuisance functions.
In the next section we formally define the causal parameters of interest using the potential outcome framework \citep{rubin1974estimating}, and introduce the assumptions yielding identification, and locally efficient and uniformly valid inference when using \ac{AIPW} estimators. We also introduce the convolutional network architectures with which we propose to fit the nuisance functions used by \ac{AIPW}, followed by our main theoretical results, including conditions to obtain uniformly valid inference when using \ac{CNN} based \ac{AIPW} estimation. Section 3 presents numerical experiments illustrating the finite sample behaviour of this estimation strategy under different data generating mechanisms. The proposed estimator is compared to \ac{AIPW}, \ac{TMLE} \citep{van2011targeted} and \ac{OR} estimation \citep{Tan:2007,moosavi2021costs} using 
% \ac{MLP} 
fully-connected feed-forward \ac{ReLU} based neural networks (\ac{MLP} in \citeauthor{farrell2018deep}, \citeyear{farrell2018deep}) and Lasso to fit the nuisance functions. In Section 4, we study the effect of early retirement (at 62 years old), compared to retiring later in life, on morbidity and mortality outcomes \citep{doi:10.1177/0049124117729697}. We use population wide Swedish register data and follow cohorts born in 1946 and 1947 for which we have a rich reservoir of potential pre-treatment confounders, including hospitalization and income histories. \ac{CNN} allows us to consider that such life histories may contain location-invariant patterns that confound the causal effects of the treatment (decision to retire early).

\section{Theory and method}
\label{sec:meth}

\subsection{Causal parameters and uniformly valid inference}
The \ac{ACE} and \ac{ACET} of a binary treatment ($T$) are parameters defined using potential outcomes (\cite{rubin1974estimating}), respectively:
$$
\tau = \mathbb{E}(Y(1)) - \mathbb{E}(Y(0)),
$$
$$
\tau_t = \mathbb{E}(Y(1) - Y(0)|T=1),
$$
where $Y(1)$ and $Y(0)$ are the outcomes that would be observed if $T=1$ and $T=0$, respectively. For a given individual only one of these potential outcomes can be observed. This intrinsic missing data problem implies that assumptions need to be made to identify $\tau$ and $\tau_t$. For this purpose, and given a vector of observed pre-treatment covariates $X$, we assume:

\begin{assumption}
\label{assum1}
\begin{enumerate}
     \item[a.] No unmeasured confounders:
     \begin{enumerate}
      \item[(i)]   $Y(0) \indep T \mid X$.
      \item[(ii)]   $Y(1) \indep T \mid X$.
      \end{enumerate}
    \item[b.] Overlap:
     \begin{enumerate}
      \item[(i)]   $\mathbb{P}(T=0\mid X)>0$.
      \item[(ii)]   $\mathbb{P}(T=1\mid X)>0$.
      \end{enumerate}
         \item[c.] Consistency: The observed outcome is $Y=Y(1)T+Y(0)(1-T).$ 
\end{enumerate}
\end{assumption}
Note that \ac{ACET} is identified if only \ref{assum1}a(i), \ref{assum1}b(i) and \ref{assum1}c hold.
Assumption \ref{assum1}a requires that the observed vector $X$ includes all confounders. Assumption \ref{assum1}b requires that for any value $X$ both treatment levels have non-zero probability to occur, and by Assumption \ref{assum1}c one of the potential outcomes is observed for each individual, and its value is not affected by the treatment received by other individuals in the sample.

We aim at uniformly (over the class of \ac{DGP}s, for which Assumption \ref{dgp.ass} hold) valid inference while using semiparametric estimation \citep{moosavi2021costs}. Therefore, the following  \ac{AIPW} estimators of \ac{ACE} are considered
\citep{robins1994estimation,scharfstein1999rejoinder}:
$$\hat{\tau} = \mathbb{E}_n [\hat{\psi}_1(z_i) - \hat{\psi}_0(z_i)]$$
where $\mathbb{E}_n[\cdot]$ is the empirical mean operator, 
$$\hat{\psi}_t(z_i) = \frac{\mathbbm{1}\{t_i=t\}(y_i -\hat{\mu}_t(x_i))}{\hat{\mathbb{P}}[T=t|X=x_i]}+\hat{\mu}_t(x_i),$$
and $\hat{\mu}_t(x)$ is an estimator of ${\mu}_t(x)=\mathbb{E}(Y(t)\mid X=x)$.

For \ac{ACET}, we use the estimator
$$\hat{\tau}_t = \mathbb{E}_n [\hat{\psi}_{1,1}(z_i) - \hat{\psi}_{0,1}(z_i)],$$
where
$$ \hat{\psi}_{t, t^\prime}(z_i) = \frac{\hat{\mathbb{P }}[T=t^\prime |X=x_i]}{ \hat{\mathbb{P }}[T=t^\prime]}\frac{\mathbbm{1}\{t_i=t\}(y_i -\hat{\mu}_t(x_i))}{\hat{\mathbb{P}}[T=t|X=x_i]}+\frac{\mathbbm{1}\{t_i=t^\prime\}\hat{\mu}_t(x_i)}{ \hat{\mathbb{P }}[T=t^\prime]}.$$

%According to \cite{Farrell_2015}, the following assumptions are taken into account for our class of Data Generating Processes (DGPs) as we attempt to obtain validity of inference for these estimators: 
We use the following assumptions for the \ac{DGP}s.
\begin{assumption}\label{dgp.ass} We have
\begin{itemize}
    \item[a.] Let $\{(y_i,t_i,x_i),\ i=1,\ldots,n\}$ be an i.i.d. sample from $(Y,T,X)$.
    \item[b.] Let $U = Y(t) -\mu_{t}(X)$. There is some $r>0$ for which $\mathbb{E}\left(\left|\mu_{t}\left(x_{i}\right) \mu_{t^{\prime}}\left(x_{i}\right)\right|^{1+r}\right)$ and $\mathbb{E} \left(\left|u_{i}\right|^{4+r}\right)$ are bounded, for given values of $t$ and $t'$.
 \end{itemize}
\end{assumption}

The estimators of the nuisance functions have yet to be introduced for these AIPW estimation strategies. In order to get the desired results, the proposed estimators should be well behaved. More precisely, the following consistency and rate conditions are considered for the nuisance function estimators. 

\begin{assumption}\label{assum2}
Let $\hat{p}(x)$ be an estimator of ${\mathbb{P }}[T=1 |X=x_i]$ which only depends on $\{x_i,t_i\}_{i=1}^n$ \citep[assumption of ``no additional randomness",][]{Farrell_2015}. Moreover, for a given $t$ we have
\begin{itemize} 
    \item[a.]$\mathbb{E}_n[(\hat{p}(x_i) - p(x_i))^2] = o_P (1)$  and  $\mathbb{E}_n [(\hat{\mu}_t(x_i) - \mu_t (x_i))^2] = o_P(1),$
\item[b.] $\mathbb{E}_n [(\hat{\mu}_t(x_i) - \mu_t(x_i))^2]^{1/2}  \mathbb{E}_n [(\hat{p}(x_i) - p(x_i))^2]^{1/2} = o_P(n^{-1/2}),$ 
\item[c.]$\mathbb{E}_n [ (\hat{\mu}_t (x_i) - \mu_t(x_i)) (1 - \mathbbm{1} \{ t_i = t \} / \mathbb{P} [T =t | X = x_i ] ) ] = o_P (n^{-1/2})$.
\end{itemize}
\end{assumption}

The following proposition describes uniform validity results obtained by \citet[Corollary 2 and 3]{farrell2018robust} under the above regularity conditions.
\begin{prop}\label{proposition}
For each $n$, let ${\cal P}_n$
be the set of distributions obeying Assumptions \ref{assum1}a(i), \ref{assum1}b(i), \ref{assum1}c and \ref{dgp.ass}a. Further, assume Assumption \ref{dgp.ass}b holds for $t=t'=0$, and let $\hat{p}(x)$ and $\hat{\mu}_0(x)$ fulfill Assumption \ref{assum2}. Then, we have:
\begin{equation*}
    \begin{aligned}
\sup_{P \in {\cal P}_n} \left\lvert \mathbb{P}_P 
\left(\tau_t \in \left\{\hat{\tau}_t  \pm c_\alpha \sqrt{\hat{V}_t/n}\right\}\right) - \left(1-\alpha\right)\right\rvert  \rightarrow 0,
    \end{aligned}
\end{equation*}
where
$$\hat{V}_t= \frac{n^2}{n_{t}^{2}}\mathbb{E}_{n}\left[ \mathbbm{1}(t_i =1)\left(y_{i}-\hat{\mu}_{0}\left(x_{i}\right)-\hat{\tau}_t\right)^{2}\right]+  \frac{n^2}{n_{t}^{2}} \mathbb{E}_{n}\left[\frac{\hat{p}\left(x_{i}\right)^{2}}{\left(1-\hat{p}\left(x_{i}\right)\right)^{2}} \mathbbm{1}(t_i =0)\left(y_{i}-\hat{\mu}_{0}\left(x_{i}\right)\right)^{2}\right],
$$
$n_t=\Sigma_{i=1}^n \mathbbm{1}(T=t)$ and $c_\alpha = \Phi^{-1}(1 - \alpha/2)$.

Let also Assumptions \ref{assum1}a(ii), \ref{assum1}b(ii) be fulfilled and Assumption \ref{dgp.ass}b hold for $t,t' \in \{0,1\}$. Additionally, assume $\hat{\mu}_1(x)$ fulfills Assumption \ref{assum2}. Then, we have:

\begin{equation*}
    \begin{aligned}
\sup_{P \in {\cal P}_n} \left\lvert \mathbb{P}_P 
\left(\tau \in \left\{\hat{\tau} \pm c_\alpha \sqrt{\hat{V}/n}\right\}\right) - \left(1-\alpha\right)\right\rvert  \rightarrow 0,
    \end{aligned}
\end{equation*}
where
\begin{equation*}
    \begin{aligned}
\hat{V} &=  \mathbb{E}_{n}\left[\frac{\mathbbm{1}(t_i =1)\left(y_{i}-\hat{\mu}_{1}\left(x_{i}\right)\right)^{2}}{\hat{p}\left(x_{i}\right)^{2}}\right]+\mathbb{E}_{n}\left[\left(\hat{\mu}_{1}\left(x_{i}\right)- \mathbb{E}_n [\hat{\psi}_1(z_i)]\right)^2\right] \\
&+ \mathbb{E}_{n}\left[\frac{\mathbbm{1}(t_i =0)\left(y_{i}-\hat{\mu}_{0}\left(x_{i}\right)\right)^{2}}{\left(1-\hat{p}\left(x_{i}\right)\right)^{2}}\right]+\mathbb{E}_{n}\left[\left(\hat{\mu}_{0}\left(x_{i}\right)- \mathbb{E}_n [\hat{\psi}_0(z_i)]\right)^2\right].
    \end{aligned}
\end{equation*}
\end{prop}
\begin{remark}
A multiplicative rate condition as Assumption \ref{assum2}(b) is weaker than separate conditions on the two nuisance model estimators. It only requires that one of the nuisance functions is estimated at faster rate if the other one is estimated at slower rate. However, using the regularity conditions in this paper and \citet{farrell2018deep}, the rate $o_P(n^{-1/4})$ is obtained for each of the nuisance estimators separately. This is to make sure Assumption \ref{assum2}(c) for $\hat{\mu}$ is also fulfilled. 
% In other words, if we rely on Assumption \ref{assum2}(c) to get uniform valid inference  \citep[see][]{moosavi2021costs} is not fully achieved.
This assumption can be, however, dropped by considering sample splitting \citep{chernozhukov2018double}.
\end{remark}
Note, finally, that because the \ac{AIPW} estimators is based on the efficient influence function, its asymptotic variance is equal to the semiparametric efficiency bound \citep{tsiatis2006}.

\subsection{Convolutional neural networks}\label{cnn.sec}
We consider a specific \ac{CNN} architecture with parallel hidden layers structured as follows. Let the input column vector be denoted by $h^0 := (x_1, \cdots ,x_d )' \in \Omega$, where $\Omega \subseteq \mathbb{R}^d$. We consider $L$ to be the number of hidden layers in which we have $E$ number of parallel vectors. Let $\sigma$ be the \ac{ReLU} function defined on the space of real numbers as $\sigma(z) = \max(0, z)$ for $z\in \mathbb{R}$. The vectors in each hidden layer $l \in \{1, \cdots, \L\}$ have size $d_l$  and are defined by $h_e^l = \sigma( W_e^l h_e^{l-1} - b_l^e)$, where $e \in \{ 1 , \cdots, E\}$, $h^0_e=h^0$, $ b_l^e$ is a vector called bias vector in the neural network literature, 
% $\sigma$ is the \ac{ReLU} function defined on the space of real numbers as $\sigma(x) = \max(0, x)$, 
$W_e^l$ is a weight matrix defined as:

\begin{equation*}
W_e^l = 
\begin{bmatrix}
w_{0,l}^e & 0 &0 &0 & \cdots & 0 \\
w_{1,l}^e & w_{0,l}^e &0 &0 & \cdots & 0 \\
\vdots  & \ddots  & \ddots & \ddots  &  \cdots & \vdots  \\
w_{S,l}^e & w_{S-1,l}^e & \cdots & w_{0,l}^e & 0 \cdots & 0  \\
0 & w_{S,l}^e & w_{S-1,l}^e & \cdots & w_{0,l}^e  \cdots & 0 \\
\vdots  & \ddots  & \ddots & \ddots  &  \cdots & \vdots  \\
\vdots  & \cdots  & 0 & w_{S,l}^e  &  \cdots & w_{0,l}^e \\
\vdots  & \cdots  & \cdots  & 0 & w_{S,l}^e\cdots & w_{1,l}^e \\
\vdots  & \ddots  & \ddots & \ddots  & \ddots  & \vdots  \\
0  & \cdots  & \cdots & \cdots   &0 \quad  w_{S,l}^e  & w_{S-1,l}^e \\
0  & \cdots  & \cdots & \cdots   &\cdots  0   & w_{S,l}^e 
\end{bmatrix},
\end{equation*}
and the weight parameter $w_{i,l}^e$ is the $i$-the element of the filter mask $(w_{0,l}^e, \cdots, w_{S,l}^e)$. $S$ is considered fixed through all values of $l$ and $e$. We have $d_l=d_0 +Sl$, where $d_l$ is the size of vectors in the $l$-th layer and $d_0 = d$.

The structure of a \ac{CNN} as defined above provides us with the following function space on $\Omega$:
$$
\mathcal{C} = \{ \sum_{e=1}^E  c_e^{\prime} h_e^L : c_e \in \mathbb{R}^{d_L}  \}.$$

% \begin{figure}
% \centering
% \includegraphics[width=6in]{Diagram4.pdf} 
% \caption{\label{fig:1}The convolutional neural networks considered use parallel kernels of the same length $S$. Using vectors with zero-padding leads to longer vectors in the subsequent layers. In the last layer, a linear combination is used to form the output.}
% \end{figure}

Similar to \citet{zhou2020universality} (who, however, uses $E=1$) we assume that for all values of $e$ in $\{1, \cdots E\}$ and $l$ in $\{ 1, \cdots, L-1 \}$ the vector $b_l^e$ has the form $(b^e_{l,1},\allowbreak \cdots,\allowbreak b^e_{l,S},\allowbreak b^e_{l,S+1},\allowbreak \cdots,\allowbreak b^e_{l,S+1},\allowbreak b^e_{l,d_l-S+1},\allowbreak  \cdots,\allowbreak b^e_{l,d_l})'$ with the $d_l - 2S$ repeated components in the middle. Therefore, the total number of parameters (duplicate/shared weights are counted more than once), $Q$ for the neural network fulfills the following inequality:
\begin{equation}\label{orderOfQ} Q = E(d(1+s)(L-1) + s(1+s) L(L-1)/2 + (s+3)(d+sL))  \asymp EL^2,
\end{equation}
where the notation $x_n \asymp y_n$ for two sequences of random variables $x_n$ and $y_n$ means $x_n = O_P(y_n)$ and $y_n = O_P(x_n)$.

\subsection{Main results}
In this section, we provide results that allow us to use the \ac{CNN} architectures described above to fit the nuisance models needed for \ac{AIPW} estimations of \ac{ACE} and \ac{ACET}, and obtain uniformly valid inference.  In particular, we need to show that the rate conditions in Assumption~\ref{assum2} are fulfilled.

\begin{assumption}
\label{data.assum}
 Assume that $z_{i}=\left(u_{i}, v_{i}^{\prime}\right)^{\prime}, 1 \leq i \leq n$ are i.i.d. copies of $Z=(U, V)\in [-1,1]^d \times\mathcal{V}$, where $U$ is continuously distributed.  For an absolute constant $M>0$ and a target $f^{*}$ (a
nuisance function),
 % assume $\left\|f^{*}\right\|_{\infty} \leq M$ 
 assume $\mathcal{V} \in [-M, M]$ and $\left\|f^{*}\right\|_{\infty} \leq M$.
 \end{assumption}

% Let us introduce some notation. Consider $z=\{u_i, v_i \}_{i=1}^{n}$. Moreover,

\begin{assumption}\label{loss.assum} Let $\ell$ be a loss function that, for any arbitrary functions $f$ and $g$ and $z$ a realization of $Z$, fulfills:
\begin{equation*}
    \begin{split}
        & |\ell(f,z)- \ell(g,z)| \leq C_\ell |f(x) - g(x)|, \\
        & c_1 \mathbb{E}((f - \tilde{f})^2) \leq \mathbb{E}(\ell(f,Z)) - \mathbb{E}(\ell(\tilde{f},Z)) \leq c_2 \mathbb{E}((f - \tilde{f})^2),
    \end{split}
\end{equation*}
where $\tilde{f} = \argmin \mathbb{E} (\ell(f,Z))$. 
\end{assumption}
Further, let $\mathcal{F}$ be an arbitrary set of functions. 
% For a loss function with the above properties, we
We define: 
\begin{equation}\label{fhatdefinition}
    \hat{f}_{\mathcal{F}} := \argmin_{\substack{f \in \mathcal{F} \\ \norm{f}_\infty \leq M^\prime }} \sum_{i=1}^{n} \ell(f,z),
\end{equation}
and
\begin{equation}\label{epsilonffprime}
    \varepsilon_{\mathcal{F},\mathcal{F}^\prime} := \sup_{\substack{f \in \mathcal{F}
    % \\ \norm{f}_\infty \leq M
    }} 
    \inf_{\substack{f^\prime \in \mathcal{F}^\prime \\ \norm{f^\prime}_\infty \leq M^\prime}} \norm{f - f^\prime}_\infty ,
\end{equation}
where $\norm{\cdot}_\infty$ is the supremum norm.
The following result is a special case of Theorem 2(b) given in \citet{farrell2018deep}, and gives the rate of convergence for our estimate  $\hat{f}_{\mathcal{C}}$ of a target $f^*$ (a nuisance function). 
Prior to presenting the theorem, the target space in which we seek to find the nuisance functions need to be defined.
We use the notation $W^{\beta, \infty}(\Omega)$ for the Sobolev space with smoothness $\beta \in \mathbb{N}_+$. 
We consider $\mathcal{W}$ to be the unit ball on the Sobolev space defined on $[-1,1]^d$; $\mathcal{W}  := B_1( W^{\beta, \infty}([-1,1]^d))$ such as:
$$\mathcal{W} := \{  f:   \max_{\substack{\alpha, |\alpha| \leq \beta}} \norm{D^\alpha f(x)}_{L^\infty([-1,1]^d)} \leq 1  \},$$
where $\norm{f}_{L^\infty(\Omega)}$ is the essential supremum norm of the absolute value of a function $f$ defined on $\Omega$,  $\alpha = (\alpha_1, \dotsm , \alpha_d)^\prime, |\alpha| = \alpha_1 + \dotsm + \alpha_d$, and $D^\alpha f$ is the weak derivative.

\begin{theorem}\label{theoremfarrell}
% Assume that $z_i =  (u_i^\prime, v_i)^\prime$, $i=1,\ldots,n$, are i.i.d. copies of $Z= (U,V) \in [-1,1]^d \times\mathcal{V}$, where $U$ is continuously distributed. Further, 
Let $f^* \in \mathcal{W}$, and assume that Assumptions \ref{data.assum}-\ref{loss.assum} hold. For an absolute constant $M > 0$ and $M^\prime= 2M$, let the approximation error $\varepsilon_{\mathcal{W},\mathcal{C}}$ be defined as in \eqref{epsilonffprime}.
% and let $\mathcal{V} \in [-M,M]$. 
With probability at least $1- e^{-\gamma}$, and for large enough $n$,
\begin{equation}\label{farrellbound}
    \begin{split}
       \mathbb{E}_n [(\hat{f}_{\mathcal{C}} - f^*)^2] \leq C \Bigg( \frac{QL\log Q}{n} \log n +\frac{\log \log n + \gamma}{n} +\varepsilon^2_{\mathcal{W},\mathcal{C}} \Bigg),
    \end{split}
\end{equation}
where the constant $C>0$ is independent of $n$, but may depend on $d$, $M$, and other fixed constants.
\end{theorem}
%\begin{assumption}\label{sampleassumption}
 
%\end{assumption}
 Our next step is to bound the term $\varepsilon^2_{\mathcal{W},\mathcal{C}}$ in the above inequality on the estimation error.
%  Now, we give a result for bounding the term $\varepsilon^2_{\mathcal{W},\mathcal{C}}$ in the above inequality on the estimation error. 
 The following result is similar to Theorem 1 in \cite{zhou2020universality}. However, we allow here the number of parallel layers to grow with $n$, which translates into faster growing function space for our specific \ac{CNN} architectures,  thereby a larger decay rate for the error.
\begin{theorem}\label{theoremepsiloncnn}
Let $2 \leq s \leq d$. If $L \geq 2d/(s-1)$ and $f \in \mathcal{W}$, $\norm{f}_\infty \leq M$ and an integer index $\beta > 2+ d/2$, then there exist $w, b$ and $f_{L,E}^{w,b} \in \mathcal{C}$ such that
$$ \norm{f - f_{L,E}^{w,b}}_{C([-1,1]^d)} \leq c 2^{d/2} \sqrt{\log (LE)} \frac{1}{LE}^{\frac{1}{2} + \frac{1}{d}}, $$
where $c$ is an absolute constant and therefore
$$
  \varepsilon_{\mathcal{W},\mathcal{C}} \leq C \sqrt{\log (LE)} \frac{1}{LE}^{\frac{1}{2} + \frac{1}{d}}.
$$
\end{theorem}

%we have three conditions for beta j s and d in theorem one. in the corollary 1 we have a rate condition for J and the other three should fulfill the conditions 

% \begin{corollary}\label{corollary1} Let $E \asymp n^{\frac{d}{2d+4}}$ and $J \asymp n^{\frac{1}{4d+8}} \log^2(n)$. Moreover, let $\beta$, $s$ and $d$ fulfill the assumptions in Theorem \ref{theoremepsiloncnn}. We have 
% $$ \varepsilon_{\mathcal{W},\mathcal{C}} \asymp \frac{1}{n^{\frac{d+1}{4d}} \log^{\frac{d+4}{2d}}n}$$
% \end{corollary}
% \begin{proof}
% Using Theorem \ref{theoremepsiloncnn} for any $f \in B_1(W^{\beta, \infty}([-1,1]^d)$ there exist $w, b$ and $f_{J,E}^{w,b} \in \mathcal{C}$ such that
% $$\norm{f - f_{J,E}^{w,b}}_{C([-1,1]^d)} \leq c 2^{d/2} \frac{\sqrt{ \frac{d+1}{2d+4} \log n + \log \log^2 n } }{ n^{\frac{d+1}{4d}} \log^{\frac{d+2}{d}}n  } $$
% Therefore we have
% $$ \varepsilon_{\mathcal{W},\mathcal{C}} \leq c 2^{d/2} \frac{\sqrt{ \frac{d+1}{2d+4} \log n + \log \log^2 n } }{ n^{\frac{d+1}{4d}} \log^{\frac{d+2}{d}}n  } \asymp \frac{1}{n^{\frac{d+1}{4d}} \log^{\frac{d+4}{2d}}n}$$
% \end{proof}

Finally, our main result gives conditions for the \ac{CNN} nuisance function fit to fulfill Assumption \ref{assum2}, and therefore conditions for \ac{AIPW} estimation based on \ac{CNN} fit to yield uniformly valid inference for \ac{ACE} and \ac{ACET} (Proposition \ref{proposition}).

\begin{theorem}\label{theoremrates}
% Let the conditions in Theorem \ref{theoremfarrell} hold for $u_i=x_i$, $v_i=y_i$ or $t_i$, and $f^*=\mu_t$ or $p$.
Let the conditions in Theorem \ref{theoremfarrell} hold for $u_i=x_i$, $v_i=y_i$, and $f^*=\mu_t$. Moreover the same conditions are fulfilled for $u_i=x_i$, $v_i=t_i$, and $f^*=p$.
Let also $E \asymp n^{\frac{d}{2d+4}}$, $L \asymp n^{\frac{1}{4d+8}} \log^2(n)$, $\beta=\beta_p \wedge \beta_\mu$, and $s$ and $d$ fulfill the assumptions of Theorem \ref{theoremepsiloncnn}. Then, the rate conditions in Assumption \ref{assum2} are fulfilled for nuisance functions estimators  (\ref{fhatdefinition})
 with $\cal F=\cal C$.

%($\hat{p} = \hat{p}_{\mathcal{C}}$ and $\hat{\mu_t} = 
% \hat{\mu}_{t,\mathcal{C}}$.
%let the assumptions of Corollary \ref{corollary1} hold.
% \\
% (a) $\mathbb{E}_n[(\hat{p}(x_i) - p(x_i))^2] = o_P (1)$  and  $\mathbb{E}_n [(\hat{\mu}_t(x_i) - \mu_t (x_i))^2] = o_P(1),$
% \\
% (b) $\mathbb{E}_n [(\hat{\mu}_t(x_i) - \mu_t(x_i))^2]^{1/2}  \mathbb{E}_n [(\hat{p}(x_i) - p(x_i))^2]^{1/2} = o_P(n^{-1/2}),$ and
% \\
% (c)  $\mathbb{E}_n [ (\hat{\mu}_t (x_i) - \mu_t(x_i)) (1 - \mathbbm{1} \{ t_i = t \} / \mathbb{P} [T =t | X = x_i ] ) ] = o_P (n^{-1/2})$\\
% where $t \in \{0,1\}$.
\end{theorem}

\begin{remark}
The rate result for \ac{CNN} based estimators in Theorem \ref{theoremrates} corresponds to similar results for 
% fully-connected feedforward \ac{ReLU} based estimators in 
\ac{MLP} in \citet[Theorem 3]{farrell2018deep}. However, the conditions on smoothness is less strict in our case. The rate of growth for the number of parameters considered for \ac{MLP} based estimators in \citet[Theorem 1]{farrell2018deep} is $n^\frac{d}{\beta +d}\log^5 n$, while we have considered the growth rate $n^\frac{d+1}{2d+4}\log^4(n)$. These rates are similar up to a log factor if $d$ is large and $\beta$ is close to $d$. Notice, however, that these are not necessarily minimal rates required for valid inference on \ac{ACE}(T). 
\end{remark}
% \section{?Choosing parameters of CNN structure}

% First suggestion is to use the algorithm in \citet{cui2020selective} 
% \begin{figure}
%     \centering
%     \includegraphics{Capture.PNG}
%     \label{fig:my_label}
% \end{figure}

\section{Simulation Study}
\label{sec:simu}

We perform a simulation study to evaluate the use of the \ac{CNN} together with \ac{AIPW} estimation strategy of average causal effects  proposed above. We focus here on $\tau_t$ (\ac{ACET}) but similar results are obtained for $\tau$. Comparisons are made with other algorithms to fit nuisance functions, including, 
% fully-connected neural networks (\ac{MLP}
\ac{MLP}, \citep{farrell2018deep} and post-lasso estimators \citep{Farrell_2015}. Moreover, we also include alternative strategies to \ac{AIPW}, including \ac{OR} estimation \citep{Tan:2007} and targeted learning \citep{van2011targeted}.

The simulation results are based on 1000 replications.  The neural networks initial weights are different for each replication, which makes the results more robust.
%We consider three designs using a third order polynomial and some piece-wise constant functions whose value depend on whether values of a subseqence of the time series follows a certain pattern.
%which are of subsets of four dimensional spaces, and 3)a discontinuous function that contains uncountable number of jumps in the zero neighborhood intervals. Half of the simulations use sequences of length five as the covariates and the other two one use time-series of length ten. 
%We use a time series of size ten in the first two setting and a time series of size 5 in the third one. 
% Different DGPs are considered, where first a
A sequence of ten covariates are independently generated having normal distributions with increasing means and varying variances: $X_1 \sim N(100,20)$, $X_2 \sim N(102,15)$, $X_3 \sim N(105,13)$, $X_4 \sim N(107,11)$, $X_5 \sim N(109,8)$, $X_6 \sim N(110,20)$, $X_7 \sim N(112,15)$, $X_8 \sim N(115,13)$, $X_9 \sim N(117,11)$, and $X_{10} \sim N(119,8)$. While these variables are independent, their ordering is going to be important in how the outcomes and treatment assignment are generated as described below in two different settings. 
% Error terms for the outcome models, $e_1$ and $e_0$, are generated from the standard normal distribution. Three distinct settings are considered with increasing complexity, where different type of patterns in the sequence of $X_t, t=1,\ldots,10$, are used in the generation of the potential outcomes and the treatment assignment:
\begin{description}
    \item[Setting 1] Potential outcomes and treatment assignment are generated as follows:
    % using polynomial functions of the difference of some neighboring variables in the sequence:
    $$Y(0) = 1+0.001( (X_2-X_1)^2+ (X_4-X_3)^3+(X_6-X_5)^2+(X_8-X_7)^3 +(X_{10}-X_9)^2) + e_0,$$ 
$$Y(1) = 2-0.001((X_2-X_1)^2+ (X_4-X_3)^3+(X_6-X_5)^2+(X_8-X_7)^3 +(X_{10}-X_9)^2) + e_1,$$
\begin{flalign*}
\mathbb{P}(T=1|X) = &1/(1+exp(5\times10^{-6}((X_2-X_1)^2+ &\\
& (X_4-X_3)^3+(X_6-X_5)^2+(X_8-X_7)^3 +(X_{10}-X_9)^2))).&
\end{flalign*}
In this setting, 
% polynomial functions are used 
polynomial functions of the difference of some neighboring variables in the sequence are used
to generate the nuisance models.
% The linear regressors in post-lasso estimation are allowed to consider up to power three polynomials. Thus, linear models should perform better than flexible estimators in this simulation.
\item[Setting 2] Potential outcomes and treatment assignment are generated as follows: 
$$Y(0) = 1+(l_1(X_1, X_2, X_3, X_4)+l_2(X_4, X_5, X_6, X_7)+ l_3(X_6, X_7, X_8, X_9))+ e_0, $$
$$Y(1) = 2-(l_1(X_1, X_2, X_3, X_4)+l_2(X_4, X_5, X_6, X_7)+ l_3(X_6, X_7, X_8, X_9))+ e_1,$$
\begin{flalign*}
        \mathbb{P}(T=1|X) = & 1/(1+exp(0.05 X_5 - &\\ 
        &0.1(l_1(X_1, X_2, X_3, X_4)+l_2(X_4, X_5, X_6, X_7)+ l_3(X_6, X_7, X_8, X_9)))),&
\end{flalign*}
    where
    \begin{equation*}
l_1 (x,y,z,w)=
    \begin{cases}
    10,&   \text{if } y/x,\ z/y, \text{ and }w/z>1.15, \\ 
    0,&   \text{otherwise},
    \end{cases}
\end{equation*}

\begin{equation*}
l_2 (x,y,z,w)=
    \begin{cases}
     5,&   \text{if } y/x,\ z/y, \text{ and }w/z<1.05, \\ 
     0,&   \text{otherwise},
    \end{cases}
\end{equation*}

\begin{equation*}
l_3 (x,y,z,w)=
    \begin{cases}
      3,&   \text{if } (y-1.1x)×(z-1.1y)×(w-1.1z)<0, \\ 
    0,&   \text{otherwise}.
    \end{cases}
\end{equation*}
 Here, the output of $l_1$ and $l_2$ are non-zero only if an increase by a factor bigger than 
1.15 and a decrease by a factor bigger than 1.05 is found in all consecutive pair of covariates in the input of size four, respectively. The value of $l_3$ is nonzero if an increase by a factor bigger
than 1.1 is followed by a decrease and followed by another increase by a factor of bigger than 1.1 or if the oscillation has the opposite direction.

\end{description}

\begin{figure}
\centering
\includegraphics[width=6in]{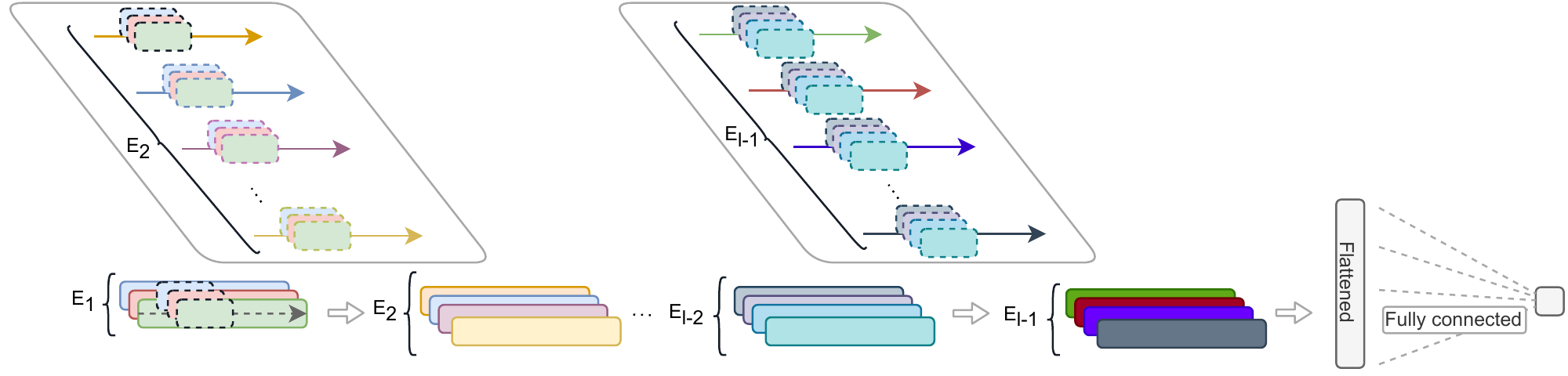} 
\caption{\label{fig:2}Illustration of the convolutional neural network utilized in the simulations and the real data study.
% It contains several feature maps (channel) in each layer. Forming each feature map in the next layers need many one-dimensional kernels (convolution filters). Each kernel is applied to exactly one of the previous layer channels, and the results are summed together to create the new feature map in the next layer.
Each layer $l$ contains different number of channels $E_l$, where $E_0$ (not shown in the Figure) is the number of time series in the set of input covariates. Each channel (feature map) in layer $l$ (e.g. the blue vector in the second layer) is formed by first applying $E_{l-1}$ number of kernels (convolution filters) on the vectors in layer $l-1$ (as demonstrated by the blue arrow in the set of kernels applied on the first layer) and then summing the results. There exist some covariates which are not time series. Inclusion of those vectors in the neural network is not illustrated in the figure for simplification purpose. A fully connected feed-forward structure has been used for those covariates and the results are then combined with the results from the time-series data in the flattening layer demonstrated above.
}
\end{figure}

The following algorithms to estimate the nuisance functions are evaluated:
\begin{description}
    
\item[CNN] A convolutional neural network using the \ac{ReLU} activation function, with architecture illustrated in Figure \ref{fig:2}. Note that this architecture is slightly more flexible than the one considered in Section~\ref{cnn.sec}, hence, we expect it to perform at least as well in terms of approximation rate. Both for the potential outcome model $\mu_0$ and the propensity score ${p}$, the number of channels in the input layer is one. For the outcome models 128 channels in the first hidden layer and 16 channels in the second hidden layer have been used, while 32 channels in the first hidden layer and 8 channels in the second hidden layer have been used for fitting the propensity score. We have implemented the \ac{CNN} together with \ac{AIPW} estimation strategy in the package \texttt{DNNCausal} which is available at \href{https://github.com/stat4reg/DNNCausal}{https://github.com/stat4reg/DNNCausal}.

\item[MLP] A \ac{ReLU} activation function based feedforward neural network with two hidden layers. Two different feedforward architectures are used in order to estimate $\mu_0$, and ${p}$ (propensity score). The networks for fitting the outcome model consist of 128 and 80 nodes within the two layers, respectively. For the propensity score, two layers contain 32 and 8 nodes, respectively. 

\item[post-lasso] Nuisance models are fitted in two steps, where all higher order terms up to order three are considered. First, a lasso variable selection step is performed using the R-package \texttt{hdm} (\cite{hdm}), and then maximum likelihood is used to fit the models with the selected covariates (those with coefficients not shrinked to zero by lasso). Variable selection is performed using two different strategies: double selection which takes the union of  variables selected by regressing  outcome and treatment variables and uses that to refit both models \citep{belloni2014inference,moosavi2021costs}, and single selection, which utilizes two sets of variables selected by regressing outcome and treatment, and refits each of the models using their corresponding set  \citep{Farrell_2015}. 

%\item[sl.nnet] Neural network as implemented in the R-package \texttt{Superlearner} \citep{spl}. 
\end{description}
We use the above nuisance function estimators in several different estimators of $\tau_t$, as follows:
\begin{description}
    \item[AIPW] is implemented using \ac{CNN}, \ac{MLP}, and post-lasso with both single and double selection (denoted respectively: DRcnn, DRmlp, DRss, DRds).
\item[OR] 
% Outcome regression 
estimator with post-lasso and double selection (denoted ORds).
\item[TMLE] 
% Targeted Maximum Likelihood Estimation 
using the R-package \texttt{tmle} and the function \texttt{SL.nnet} to fit all nuisance functions (denoted \ac{TMLE}).
\end{description}
Full details with code for this simulation study are available at the Github page: 

\href{https://github.com/stat4reg/Causal_CNN/blob/main/README.md}{https://github.com/stat4reg/Causal\_CNN/blob/main/README.md}.

We set sample sizes to $n=5000$ and $10000$, and run 1000 replicates. Bias, standard errors (both estimated and Monte Carlo), mean squared errors (MSE, bias squared plus variance) and empirical coverages for 95\% confidence intervals are reported in Table \ref{tab:1} and \ref{tab:2}. 

The \ac{DGP} of Setting 1 (Table \ref{tab:1}) is the least complex one  and its polynomial form is favorable to the post-lasso methods which are based on higher order terms. It is therefore expected that the post-lasso methods ORds, DRss, DRds have low bias. DRcnn has larger bias, but decreasing with $n$. All methods have similar MSEs even though DRcnn has the lowest MSE for the largest sample size. Empirical coverages are close to the nominal 95\% level for all methods. 

The \ac{DGP} of Setting 2 (Table \ref{tab:2}) is less smooth and we observe that the DRcnn and DRmlp methods have the lowest biases and MSEs, where DRcnn performs slightly better. \ac{TMLE} and post-lasso both have such a large bias that the confidence interval coverages are far from being at the nominal level. It should be noted, however, that the \ac{TMLE} uses the built-in feedforward neural network, which is not adjusted to the generated datasets, and so the comparison is not a fair one between \ac{TMLE} and the DR estimators.

\npdecimalsign{.}
\nprounddigits{3}

\begin{table}\caption{\label{tab:1}Setting 1 \ac{DGP}: Bias, standard errors (both estimated, Est sd, and Monte Carlo, MC sd), MSEs and empirical coverages for 95\% confidence intervals over 1000 replicates.}
    \centering 
    \begin{tabular}{ln{2}{3} n{2}{3} n{2}{3} n{2}{3} n{2}{3}}
\hline
   & \multicolumn{1}{c}{bias} & \multicolumn{1}{c}{coverage} & \multicolumn{1}{c}{MC sd} & \multicolumn{1}{c}{Est sd} & \multicolumn{1}{c}{MSE}\\
   \hline
\multicolumn{6}{c}{$n=5000$} \\
\hline\hline
DRcnn &0.184981693	&0.947	&1.378718855	&1.365854873 &1.935083908
\\

DRmlp &0.426132801	&0.94	&1.353171169	&1.331213731 &2.012661377
\\

ORds &0.000911156	&0.947	&1.403741932	&1.392029275 &1.970492242
\\

DRss &0.002306959	&0.948	&1.403853991	&1.39475808 &1.97081135
\\

DRds &0.002889993	&0.948	&1.404322827	&1.394767839 &1.972130954
\\

tmle &0.044055324	&0.944	&1.396119854	&1.384395628  & 1.951091518
\\
\hline
\multicolumn{6}{c}{$n=10000$} \\
\hline\hline
DRcnn &0.089425901	&0.958	&0.971113619	&0.971723047 & 0.951058653
\\

DRmlp &0.222762937	&0.949	&0.96079909	&0.958279168 &0.972758217
 \\

ORds &0.001219137	&0.955	&0.980652578	&0.982178154 &0.961680965
 \\

DRss &0.000693929	&0.955	&0.980953029	&0.983992944 &0.962269327
 \\

DRds &0.000701381	&0.955	&0.98097268	&0.983995167 &0.962307891
 \\

tmle &0.027570772	&0.957	&0.980766656	&0.978459917  &0.962663381
 \\

\hline
\end{tabular}
\end{table}

\begin{table}\caption{\label{tab:2}Setting 2 \ac{DGP}: Bias, standard errors (both estimated and Monte Carlo), MSEs and empirical coverages for 95\% confidence intervals over 1000 replicates.}
    \centering 
    \begin{tabular}{l n{2}{3} n{2}{3} n{2}{3} n{2}{3} n{2}{3}}
\hline
   & \multicolumn{1}{c}{bias} & \multicolumn{1}{c}{coverage} & \multicolumn{1}{c}{MC sd} & \multicolumn{1}{c}{Est sd} & \multicolumn{1}{c}{MSE}\\
   \hline
\multicolumn{6}{c}{$n=5000$} \\
\hline\hline
DRcnn &0.046948968	&0.937	&0.093829731	&0.101117953 &0.011008224
 \\

DRmlp &0.07621336	&0.89	&0.094887431	&0.099450126 &0.014812101
 \\

ORds &0.232723238	&0.206	&0.082722495	&0.085127253 &0.061003117
 \\

DRss &0.233127182	&0.26	&0.082674879	&0.0919845 &0.061183419
 \\

DRds &0.234590574	&0.25	&0.08256841	&0.091828514 &0.06185028
 \\

tmle &0.233118185	&0.204	&0.082860026	&0.085427863 &0.061209872
  \\
   \hline
\multicolumn{6}{c}{$n=10000$} \\
\hline\hline
DRcnn &0.024948577	&0.945	&0.070033594	&0.072255133 &0.005527136
 \\

DRmlp &0.036278424	&0.928	&0.070095297	&0.071613026 &0.006229475
 \\

ORds &0.224015667	&0.04	&0.061793098	&0.060581994 &0.054001406
 \\

DRss &0.224284741	&0.054	&0.061757738	&0.065293329 &0.054117663
 \\

DRds &0.226100419	&0.045	&0.061452996	&0.065213396 &0.05489787
 \\

tmle &0.224270066	&0.041	&0.061735636	&0.060738283 &0.054108351
  \\

\hline
\end{tabular}
\end{table}

\npdecimalsign{.}
\nprounddigits{2}
\npnoround

\section{Effect of early retirement on health}
\label{sec:conc}

\begin{figure}
\centering
\includegraphics[width=.49\textwidth]{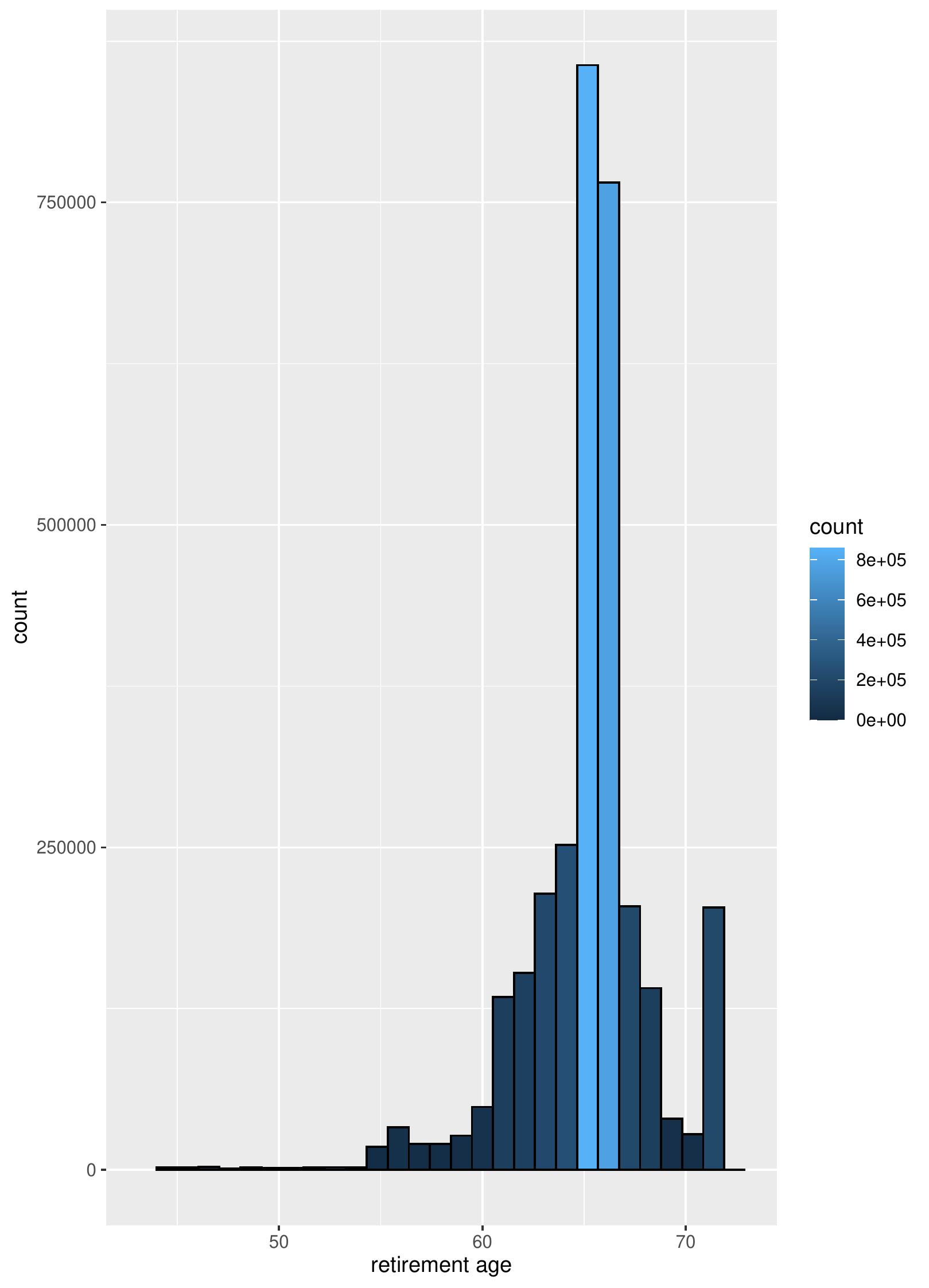} 
\includegraphics[width=.49\textwidth]{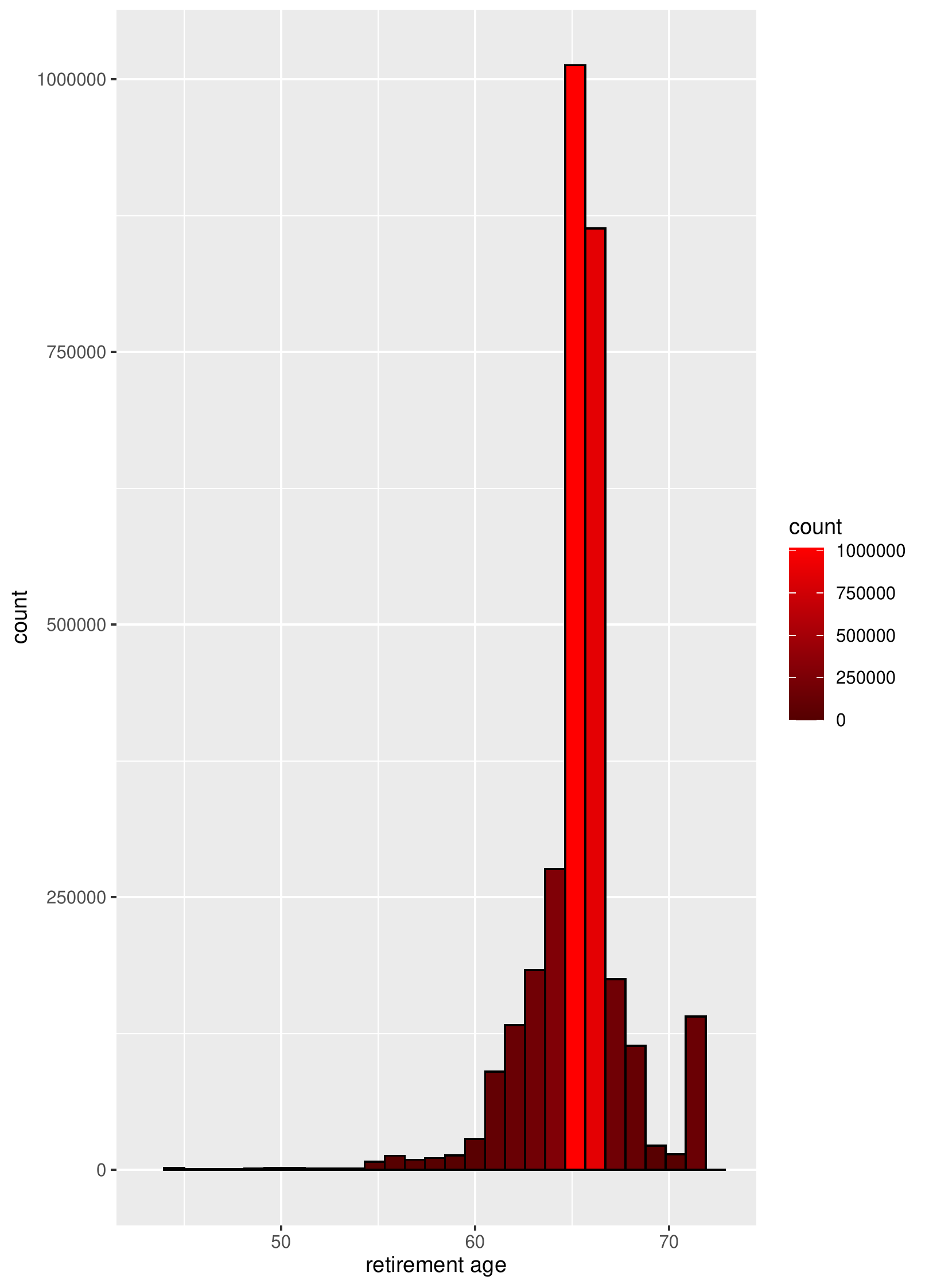} 
\caption{\label{tab:4}Retirement age for men and women who belong to the cohorts born in Sweden 1946 and 1947. The last staple in each histogram displays those who retired at age 71 or later.} \label{retirement.fig}
\end{figure}

The existing theoretical and empirical evidence on the effects of early retirement on health are mixed, and both positive and negative results may be expected and have been reported in different situations; see \citet{doi:10.1177/0049124117729697} and the references therein.
In this context, we add to the empirical evidence by studying the effect of early retirement on health using a database linking at the individual level a collection of socio-economic and health registers on the whole Swedish population \citep{simsam:2016}. This allows us to follow until 2017 two complete cohorts born in 1946 and 1947 and residing in Sweden in 1990. As health outcome we observe the number of days in hospital per year after retirement. 
To study the effects of early retirement we consider those who were still alive at age 62, and either retire at age 62 ($T=1$, treatment) or retire later ($T=0$, control group). For the cohorts studied, retirement pensions are accessible from age 61, although the usual age of retirement is 65 years of age (see Figure \ref{retirement.fig} for descriptive statistics on age of retirement). Therefore, retiring at age of 62 is considered as early retirement since it decreases your annual pension transfers compared to later retirement; see, e.g. \citet{doi:10.1177/0049124117729697} for details on the Swedish pension system. \citet{doi:10.1177/0049124117729697} also contains a study on the effects of early retirement, although based on fewer measurements for time-dependent covariates and using a matching design. This study provides new evidence by taking into account richer histories of the time-dependent covariates, and by incorporating \ac{CNN} to address complex dynamics in pre-treatment confounding covariates.

More precisely, the design of the study is as follows. An individual alive at age 62 is considered as taking early retirement at that age if hers/his pension transfers become larger than income from work at that age for the first time (i.e., they were never so earlier). For replicability, the exact definition using income and transfer variables from the Swedish registers are available at

\href{https://github.com/stat4reg/Causal_CNN/blob/main/population_description.md}{https://github.com/stat4reg/Causal\_CNN/blob/main/population\_description.md}. The health outcomes of interest are the number of hospitalization days per year during the next five years following early retirement. We, however, check first whether early retirement has an effect on survival during the five first years following early retirement. There were no (or hardly significant) such effects at the 5\% level (results reported in the appendix, Table \ref{tab:6}) and we therefore focus the analysis on the survivors when looking at effects on hospitalization. The following covariates are used as input in the \ac{CNN} architures to fit treatment and outcome nuisance models. Besides the birth year we include the the following (pre-treatment) covariates measured at 61 years of age:  marriage status, municipality, education level, Spouse education level, and the number of biological children. Moreover, we consider the measurements of the following covariates for each of the ten years preceding retirement: days of hospitalization and open health care, annual income from labour, annual income from pension, annual income from unemployment, annual income from early retirement and sickness benefit, annual compensation for illness, and spouse retirement status. Thus, covariates include eight time(-structured) series of ten observations each per individual. We do separate analyses for men and women which gives two samples of approximately 100000 individuals each.

The code giving details on the \ac{CNN} architectures and tuning parameters can be found at

\href{https://github.com/stat4reg/Causal_CNN/blob/main/population_description.md}{https://github.com/stat4reg/Causal\_CNN/blob/main/population\_description.md}. We also apply the other methods evaluated in the simulation study above.

Results are presented in Table \ref{tab:5}. 
% The naive estimation, not controlling for confounders, yields a negative effect of retiring early on health (varying within the range from 0.167 to 0.396 of number of days in average) for men and women and for the five years of follow up considered.
Based on the naive estimation not controlling for confounders, early retirement has a clear positive effect on health (varying between 0.167 and 0.396 average number of days) and for the five years of follow up considered. These naive effects are larger for women in the beginning. These negative effects disappear when controlling for the considered covariates. This is seen most clearly with \ac{CNN} which yields the effects smaller than 0.1 day (in absolute value) for all cases but one. Confidence intervals at the 95\% level cover zero in most cases. Thus, while there appeared to be a positive effect on health of early retirement, this effect was probably due to confounding.

\spacingset{0.9}

\npdecimalsign{.}
\nprounddigits{3}

\begin{table} \centering
\caption{\label{tab:5} Effect of early retirement on number of days in hospital during five years of follow up, for the early retirees, and 95\% confidence intervals.}

\begin{tabular}{r n{2}{3} n{2}{3} n{1}{3} n{2}{3} n{2}{3} n{0}{3} }
\hline
   &  \multicolumn{3}{c}{Women}                & \multicolumn{3}{c}{Men}                 \\ \hline
            \multicolumn{1}{c}{The first year}&  \multicolumn{3}{c}{n=106437}                & \multicolumn{3}{c}{n=104459}           \\ \hline \hline
naive       &-0.396361361	&\multicolumn{2}{c}{}                       &-0.167256385   &\multicolumn{2}{c}{}\\ 
DRcnn       & 0.061620095   &{(}-0.060075903{{,}}    &0.183316092{)}    &-0.036601272    &{(}-0.256368478{{,}}	&0.183165934{)}\\ 
DRmlp       & -0.061897842   &{(}-0.1923701{{,}}    &0.068574415{)}    &0.13792645    &{(}-0.078785947{{,}}	&0.354638847{)}\\ 
ORds      &-0.135237022	&{(}-0.231322569{{,}}   &-0.039151476{)}    &0.017873347	&{(}-0.155947569{{,}}	&0.191694263{)}\\ 
DRss     &-0.138313673	&{(}-0.232077559{{,}}  &-0.044549788{)}     &0.025423066	&{(}-0.148240351{{,}}	&0.199086482{)}\\ 
DRds      &-0.128306375	&{(}-0.221755045{{,}}  &-0.034857706{)}     &0.028490831	&{(}-0.14500868{{,}}	&0.201990343{)}\\ 
tmle        &-0.143151569	&{(}-0.236813917{{,}}	&-0.049489222{)}    &0.018031524	&{(}-0.155631142{{,}}	&0.191694191{)}\\ \hline
                  \multicolumn{1}{c}{The second year} &  \multicolumn{3}{c}{n=105776}                & \multicolumn{3}{c}{n=103560}          \\ \hline \hline
naive       &-0.353950676   &\multicolumn{2}{c}{}                       &-0.192061728   &\multicolumn{2}{c}{}\\ 
DRcnn       &-0.082153614   &{(}-0.23097387{{,}}	&0.066666642{)}     &0.121835959	&{(}-0.041655011{{,}}	&0.285326929{)}\\ 
DRmlp       & -0.083209362   &{(}-0.224066072{{,}}    &0.057647347{)}    &0.213806426    &{(}0.040710758{{,}}	&0.386902093{)}\\ 
ORds      &-0.103582902	&{(}-0.214639371{{,}}	&0.007473566{)}    &-0.00721344	&{(}-0.160728861{{,}}	&0.146301981{)}\\ 
DRss     &-0.097987398	&{(}-0.206833581{{,}}	&0.010858785{)}    &0.010702434	&{(}-0.142117382{{,}}	&0.163522249{)}\\ 
DRds      &-0.08148783	&{(}-0.190465472{{,}} 	&0.027489812{)}    &0.013820969	&{(}-0.139280738{{,}}	&0.166922677{)}\\ 
tmle        &-0.111039489	&{(}-0.219839761{{,}}	&-0.002239218{)}    &0.010660505	&{(}-0.142148868{{,}}	&0.163469877{)}\\ \hline
                  \multicolumn{1}{c}{The third year} &  \multicolumn{3}{c}{n=105039}                & \multicolumn{3}{c}{n=102547}          \\ \hline \hline
naive       &-0.277413722   &\multicolumn{2}{c}{}                       &-0.172825105   &\multicolumn{2}{c}{}\\ 
DRcnn       &0.010917574	&{(}-0.152970788{{,}}	&0.174805935{)}     &0.029932594	&{(}-0.140274345{{,}}	&0.200139534{)}\\ 
DRmlp       &0.19967292   &{(}0.029456363{{,}}    &0.369889477{)}    &1.094097028    &{(}-0.894419665{{,}}	&3.082613721{)}\\ 
ORds      &-0.03401301	&{(}-0.191795899{{,}}	&0.12376988{)}     &0.014314378	&{(}-0.136465839{{,}}	&0.165094594{)}\\ 
DRss     &-0.018106725	&{(}-0.174410203{{,}}	&0.138196754{)}    &0.024353342	&{(}-0.125933742{{,}}	&0.174640425{)}\\ 
DRds      &-0.01030773	&{(}-0.166548031{{,}}	&0.145932571{)}    &0.027449718	&{(}-0.122888525{{,}}  &0.17778796{)}\\ 
tmle        &-0.017658002	&{(}-0.173959312{{,}}	&0.138643308{)}    &0.027387248	&{(}-0.122899529{{,}}	&0.177674026{)}\\ \hline
                  \multicolumn{1}{c}{The fourth year} &  \multicolumn{3}{c}{n=104293}                & \multicolumn{3}{c}{n=101502}          \\ \hline \hline
naive       &-0.178659174   &\multicolumn{2}{c}{}                       &-0.215792345   &\multicolumn{2}{c}{}\\ 
DRcnn       &0.067288162	&{(}-0.180808041{{,}}	&0.315384364{)}    &0.084850885 	&{(}-0.082353649{{,}}	&0.252055419{)}\\ 
DRmlp       &0.804122757   &{(}-0.431518756{{,}}    &2.039764271{)}    &-0.517327941    &{(}-1.722586428{{,}}	&0.687930546{)}\\ 
ORds      &0.032356196	&{(}-0.185784779{{,}}	&0.250497171{)}    &-0.031857133	&{(}-0.194441892{{,}}	&0.130727625{)}\\ 
DRss     &0.041811625	&{(}-0.175812052{{,}}	&0.259435302{)}    &-0.015765319	&{(}-0.177808748{{,}}  &0.14627811{)}\\ 
DRds      &0.055738373	&{(}-0.161956447{{,}}   &0.273433194{)}    &-0.008338728	&{(}-0.170409473{{,}}	&0.153732017{)}\\ 
tmle        &0.032982071	&{(}-0.184638486{{,}}	&0.250602629{)}    &-0.014729201	&{(}-0.176770323{{,}}  &0.14731192{)}\\ \hline
                  \multicolumn{1}{c}{The fifth year}  &  \multicolumn{3}{c}{n=103477}                & \multicolumn{3}{c}{n=100325}         \\ \hline \hline
naive       &-0.231905073   &\multicolumn{2}{c}{}                       &-0.22953711	&\multicolumn{2}{c}{} \\ 
DRcnn       &-0.011610013	&{(}-0.184664749{{,}}	&0.161444724{)}    &-0.095444006	&{(}-0.27603619{{,}}	&0.085148178{)}\\ 
DRmlp       &-0.978913388   &{(}-2.981803548{{,}}    &1.023976771{)}    &-0.152618497    &{(}-0.442810181{{,}}	&0.137573187{)}\\ 
ORds      &-0.018519994	&{(}-0.169105975{{,}}	&0.132065987{)}    &-0.049856024	&{(}-0.194599047{{,}}	&0.094886999{)}\\ 
DRss     &-0.013192573	&{(}-0.163152898{{,}}	&0.136767753{)}    &-0.035395015	&{(}-0.179461239{{,}}	&0.108671209{)}\\ 
DRds      &0.00020757	    &{(}-0.150223281{{,}}	&0.150638421{)}    &-0.028528928	&{(}-0.172649411{{,}}	&0.115591555{)}\\ 
tmle        &-0.013794288	&{(}-0.163755286{{,}}	&0.13616671{)}     &-0.034034271	&{(}-0.178092483{{,}}	&0.110023941{)}\\    \hline  
\end{tabular}
\end{table}

\spacingset{1}

\npnoround

\section{Discussion}
We have proposed and studied a semiparametric estimation strategy for average causal effects which combine convolutional neural network with \ac{AIPW}. As long as the conditions given are met, this strategy yields locally efficient and uniformly valid inference. The use of \ac{CNN} has the advantage over fully-connected feed forward neural networks that they are more efficient on the number of weights (free parameters) used for approximating the nuisance functions, and are geared to take into account time invariant local features in the data. 

A main contribution of the paper is to show under which conditions \ac{CNN} fits of  nuisance functions achieve $n^{-1/4}$ convergence rate required to obtain uniformly valid inference on an \ac{ACE}.  Using the result in \citet{farrell2018deep} in which convergence rate of a \ac{ReLU}-based feed-forward network is shown to follow Equation \eqref{convergencebound}, we show that the rate conditions in Assumption \ref{assum2} are fulfilled. Specifically, we use \ac{CNN} with a given complexity for which we show that approximation rate in \eqref{convergencebound} is small enough.
 A key component of result in \citet{farrell2018deep} is the upper bound on empirical process terms found in \citet{bartlett2005local} based on Rademacher averages, which are measures of function space complexity. Global Rademacher averages do not provide fast rates as in \eqref{convergencebound}. To derive this fast rate, localization analysis is employed which takes into consideration only intersection of the function space with a ball around the goal function; considering that in reality the algorithm only searches in a neighborhood around the true function. Moreover, the tight bound on Pseudo-dimension of deep nets found in \citet{bartlett2019nearly} is used. 
%  The fact that the rate is fast and the complexity-independent coefficient of the first term in \eqref{convergencebound} is small allows us to select a complexity large enough for our network and thus get a small approximation rate while the first term remains small.

We also present numerical experiments showing the performance of the estimation strategy in finite sample and compare it to other machine learning based estimation strategies. The applicability of the method is illustrated through a population wide observational study of the effect of early retirement on hospitalization in Sweden.

\section{Appendix}

In this section we also consider the Sobolev space $H^\beta(\Omega)$ that is defined using the $L^2$ norm instead of the $L^\infty$ norm. Moreover, let $\norm{f}_{L^2(\Omega)}$ be the $L^2$ norm of a function $f$ defined on $\Omega$.  
 
\subsection{Proof of Theorem 2}
\begin{proof}

Let $\Omega = [-1,1]^d$. For any $\alpha$ that satisfies $|\alpha| \leq \beta$, by Hölder's inequality, we have:

%$$ \norm{f}_{L^p(X)} \leq \mu (X)^{1/p-1/q}\norm{f}_{L^q(X)} $$
%then for $p=2$ and $q = \infty$ we have:
$$ \norm{D^\alpha f}_{L^2(\Omega)} \leq 2^{d/2}\norm{D^\alpha f}_{L^\infty(\Omega)}. $$
Therefore, $f \in H^\beta(\Omega)$ and $\norm{f}_{H^\beta(\Omega)} \leq 2^{d/2}$ hold by $\max_{\substack{\alpha, |\alpha| \leq \beta}} \norm{D^\alpha f}_{L^\infty(\Omega)} \leq 1$. %\begin{equation*}
%\begin{split}
%\max_{\substack{\alpha, |\alpha| \leq \beta}} \esssup_{\substack{x \in \Omega}} |D^\alpha f(x)| \leq 1  &\Rightarrow \max_{\substack{\alpha, |\alpha| \leq \beta}} \norm{D^\alpha f}_{L^2(\Omega)} \leq 2^{d/2}
%\\
%&\Rightarrow f \in B_{2^{d/2}} (H^\beta(\Omega))
%\end{split}
%\end{equation*}
 Moreover, $\Omega$ has a Lipschitz domain interior and measure zero boundary. Therefore, by Sobolev extension theorem \cite[Theorem 5]{10.2307/j.ctt1bpmb07} there is an extension $F \in H^\beta(\mathbb{R}^d)$ of $f$ for which 
$$\norm{F}_{H^\beta(\mathbb{R}^d)} \leq C \norm{f}_{H^\beta(\Omega)} \leq C 2^{d/2},$$
where $C$ is a constant.

By $F \in H^\beta(\mathbb{R}^d)$ and the result in \citet[Theorem 2]{klusowski2018approximation}, we have 
$$
\norm{F-F_{m}}_{C(\Omega)} \leq c_0 v_{F,2} \max \{ \sqrt{\log m}, \sqrt{d}\} m^{-\frac{1}{2}-\frac{1}{d}},
$$ 
 where $m = E \floor*{\frac{(s-1)L}{d}-1 }$,
 %$m_0 =  \floor*{\frac{(s-1)L}{d}-1}$, 
$
    F_{m}(x) = \beta_0 + \alpha_0  x + \frac{v}{m} \sum_{k=1}^{m} \beta_k(\alpha_k x - t_k)_+,
$
$\beta_k \in [-1,1]$, $\norm{\alpha_k}_1 =1$, $t_k \in [0,1]$, $\beta_0 = F(0)$, $\alpha_0 = \nabla F(0)$ and $|v| \leq 2v_{F,2}$. Here, $v_{F,2} := \int_{\mathbb{R}^d} \norm{\omega}^2_1 |\hat{F}(\omega)|d\omega \leq c_{d,\beta} \norm{F}_{H^\beta(\mathbb{R}^d)}$, where $c_{d,\beta}$ is the finite constant $\norm{\norm{\omega}_1^2 (1+|\omega|^2)^{-\beta/2}}_{L^2}$ and $\hat{F}(\omega)$ is the Fourier transform of $F$.

Let $F_{m}^i := \frac{1}{E}\beta_0 + \frac{1}{E} \alpha_0  x + \frac{v}{m} \sum_{k=im +1}^{(i+1)m} \beta_k(\alpha_k x - t_k)_+$. It is shown in \cite{zhou2020universality}, using $s \geq 2$ and $L \geq 2d/(s-1)$, that $F_{m}^i|_\Omega$ belongs to the following function space
$$
\mathcal{C}^{w,b}_{L} = \left\{  \sum_{k=1}^{d_L} c_k h_k^{(L)} (x) : c \in \mathbb{R}^{d_L}  \right\}.$$
Therefore, by the definition of $\mathcal{C}$, we can directly conclude that $F_{m}|_\Omega$ belongs to $\mathcal{C}$. Let $f_{L,E}^{w,b} := F_m|_\Omega $. Then, we have 
\begin{equation}
    \begin{split}
        \norm{f - f_{L,E}^{w,b}}_{C(\Omega)} &= \norm{F - F_m}_{C(\Omega)}\\
        &\leq c_0 v_{F,2} \max \{ \sqrt{\log m}, \sqrt{d}\} m^{-\frac{1}{2}-\frac{1}{d}} \\
        &\leq c_0 c_{d,\beta} \norm{F}_{H^\beta(\mathbb{R}^d)}  \max \{\sqrt{\log m}, \sqrt{d}\} m^{-\frac{1}{2}-\frac{1}{d}}.
    \end{split}
\end{equation}
The final result is obtained from the facts that $\frac{1}{2}(s-1)LE \leq md \leq (s-1)LE$, $s \geq 2$ and $\beta > 2+ d/2$ ($c_{d,\beta}$ is bounded).

\end{proof}
\subsection{Proof of Theorem 3}

\begin{proof}
The rate conditions (a) and (b) of Assumption \ref{assum2} are fulfilled if both $\mathbb{E}_n [(\hat{\mu}_t(x_i) - \mu_t(x_i))^2]$ and $\mathbb{E}_n [(\hat{p}(x_i) - p(x_i))^2]$ are $o_P(n^{-1/2})$. For this, the three terms in the upper bound \eqref{farrellbound} in Theorem \ref{theoremfarrell} must be $o_P(n^{-1/2})$. For the first term, based on \ref{orderOfQ}, we have
\begin{align*}
    \frac{QL\log Q}{n} \log n 
%     & =
% \frac{E(d(1+s)(1-L) + s(1+s) L(L-1)/2 + (s+3)(d+sL))L\log (E(d(1+s)(1-L) + s(1+s) L(L-1)/2 + (s+3)(d+sL))}{n} \log n\\
&\asymp  \frac{EL^3 \log (EL^2)}{n} \log n \asymp\frac{  \log^6(n) \log ( n^{\frac{d+1}{2d+4}} \log^4n)}{n^{\frac{2d+5}{4d+8}}} \log n = o_P(n^{-1/2}).
\end{align*}
The second term is also $o_P(n^{-1/2})$ if $\gamma = o(n^{1/2})$. For the third term, using Theorem \ref{theoremepsiloncnn} we have
\begin{align*}
\varepsilon_{\mathcal{W},\mathcal{C}}\leq \frac{\sqrt{ \frac{d+1}{2d+4} \log n + \log \log^2 n } }{ n^{\frac{d+1}{4d}} \log^{\frac{d+2}{d}}n  }. 
\end{align*}
To prove condition (c) of Assumption \ref{assum2}, we use the proof of lemma \citet[Lemma 10]{farrell2018deep}. In this proof it is shown that for an arbitrary class of feedforward neural networks with probability at least $1 - \exp(-n^{\frac{d+3}{4d+8}}\log^6n)$
\begin{equation*}
    \begin{split}
 \mathbb{E}_n \Bigg[ (\hat{\mu}_{t,\mathcal{C}} (x_i) - \mu_t(x_i)) \Bigg( 1 - \frac{\mathbbm{1}\{t_i=t\}}{\mathbb{P}[T=t|X=x_i]} \Bigg) \Bigg]  \leq & C \Bigg(\frac{QL\log Q}{n}\log n + \\
 &\frac{\log \log n + n^{\frac{d+3}{4d+8}}\log^6n }{n} + \varepsilon^2_{\mathcal{W},\mathcal{C}}  \Bigg).
    \end{split}
\end{equation*}
Therefore, we can conclude that under the stated conditions
\begin{equation*}
    \begin{split}
        &\mathbb{E}_n \Bigg[ (\hat{\mu}_t (x_i) - \mu_t(x_i)) \Bigg( 1 - \frac{\mathbbm{1}\{t_i=t\}}{\mathbb{P}[T=t|X=x_i]} \Bigg) \Bigg] \leq C \mathbb{E}_n [(\hat{\mu}_t(x_i) - \mu_t(x_i))^2] = o_P(n^{-1/2}).
    \end{split}
\end{equation*}
\end{proof}

\subsection{Early retirement effects on survival}
Additional results on the effect of early retirement on survival can be found in Table \ref{tab:6}.

\npdecimalsign{.}
\nprounddigits{3}

\spacingset{0.9}
\begin{table}[!h]
\caption{\label{tab:6} Effect of early retirement on survival (binary outcomes death=1), during five years of follow up for the early retirees, and 95\% confidence intervals.}
    \centering
    \begin{tabular}{r n{2}{3} n{2}{3} n{1}{3} n{2}{3} n{2}{3} n{0}{3} }
    \hline
          &  \multicolumn{3}{c}{Women n=106916}                & \multicolumn{3}{c}{Men n=105149}                 \\ \hline
                   \multicolumn{7}{c}{The first year after treatment is considered for the outcome}           \\ \hline \hline
        naive       & 0.002207142   & \multicolumn{2}{c}{}  & -0.00142164 &\multicolumn{2}{c}{}  \\
        DRcnn       & 0.00485995    &{(} 0.001732862{{,}}   & 0.007987039{)}  & -0.001058343    &{(} -0.005274609{{,}}  & 0.003157924{)} \\ 
        DRmlp       & 0.003806873 &{(} 0.000395266{{,}} & 0.007218479{)} & -0.003162403 &{(} -0.012943554{{,}} & 0.006618749{)} \\ 
        ORds      & 0.003328143   &{(} -0.193836569{{,}}  & 0.200492854{)}  & 0.00026753      &{(} -0.150081257{{,}}  & 0.150616317{)} \\ 
        DRss     & 0.003392179   &{(} 0.000379314{{,}}   & 0.006405043{)}  & 0.000290068     &{(} -0.00211477 {{,}}  & 0.002694906{)} \\ 
        DRds      & 0.003492589   &{(} 0.00048229{{,}}    & 0.006502889{)}  & 0.000424917     &{(} -0.001980921{{,}}  & 0.002830754{)} \\
        tmle        & 0.003389389   &{(} 0.000377118{{,}}   & 0.006401661{)}  & 0.000248498     &{(} -0.00215635 {{,}}  & 0.002653345{)} \\  \hline
                    \multicolumn{7}{c}{The second year after treatment is considered for the outcome}           \\ \hline \hline
        naive       & 0.002244755   & \multicolumn{2}{c}{}  & -0.00211676 &\multicolumn{2}{c}{}  \\ 
        DRcnn       & 0.00730545    &{(} 0.002633178{{,}}   & 0.011977722{)}  & 0.001006037     &{(} -0.00474765 {{,}}  & 0.006759725{)} \\
         DRmlp       & 0.006230789 &{(} 0.001950718{{,}} & 0.01051086{)} & 0.001245953 &{(} -0.003413917{{,}} & 0.005905823{)} \\ 
        ORds      & 0.004605658   &{(} -0.16437253{{,}}   & 0.173583847{)}  & 0.001557138     &{(} -0.129855798{{,}}  & 0.132970074{)} \\
        DRss     & 0.004906075   &{(} 0.000835603{{,}}   & 0.008976548{)}  & 0.001668896     &{(} -0.001811486{{,}}  & 0.005149278{)} \\
        DRds      & 0.005434267   &{(} 0.001371078{{,}}   & 0.009497456{)}  & 0.001907049     &{(} -0.00157454 {{,}}  & 0.005388637{)} \\
        tmle        & 0.004920809   &{(} 0.000853242{{,}}   & 0.008988376{)}  & 0.001652466     &{(} -0.001827723{{,}}  & 0.005132654{)} \\ \hline
                    \multicolumn{7}{c}{The third year after treatment is considered for the outcome}           \\ \hline \hline
        naive       & 0.000305704   & \multicolumn{2}{c}{}  & -0.001668329 & \multicolumn{2}{c}{}  \\
        DRcnn       & -0.05150568   &{(} -0.166492834{{,}}  & 0.063481474{)}  & 0.005142971     &{(} 0.000127797 {{,}}  & 0.010158146{)} \\
        DRmlp       & 0.004390524 &{(} -0.000649012{{,}} & 0.00943006{)} & 0.00804965 &{(} 0.002729398{{,}} & 0.013369901{)} \\ 
        ORds      & 0.004427337   &{(} -0.149923722{{,}}  & 0.158778396{)}  & 0.004039032     &{(} -0.114198364{{,}}  & 0.122276428{)} \\
        DRss     & 0.00471048    &{(} 0{{,}}     & 0.009455949{)}  & 0.004276443     &{(} -0.000210146{{,}}  & 0.008763033{)} \\
        DRds      & 0.005309033   &{(} 0.000572128{{,}}   & 0.010045939{)}  & 0.004525989     &{(} 0    {{,}}  & 0.009013672{)} \\
        tmle        & 0.004730256   &{(} 0{{,}}     & 0.00947278 {)}  & 0.004280453     &{(} -0.000204701{{,}}  & 0.008765606{)} \\ \hline
                    \multicolumn{7}{c}{The fourth year after treatment is considered for the outcome}           \\ \hline \hline
        naive       & -0.002951911  & \multicolumn{2}{c}{} & -0.005639835 & \multicolumn{2}{c}{} \\
        DRcnn       & 0.001779057   &{(} -0.004715678{{,}}  & 0.008273791{)}  & -0.000832129    &{(} -0.006503582{{,}}  & 0.004839324{)} \\
        DRmlp       & 0.000811174 &{(} -0.004627847{{,}} & 0.006250195{)} & 0.005185165 &{(} -0.000366907{{,}} & 0.010737238{)} \\
        ORds      & 0.002705643   &{(} -0.139695932{{,}}  & 0.145107218{)}  & 0.001793698     &{(} -0.106661074{{,}}  & 0.11024847 {)}\\
        DRss     & 0.003198479   &{(} -0.002031619{{,}}  & 0.008428577{)}  & 0.002124748     &{(} -0.00288025 {{,}}  & 0.007129746{)} \\
        DRds      & 0.00383222    &{(} -0.001388967{{,}}  & 0.009053407{)}  & 0.002422892     &{(} -0.002583136{{,}}  & 0.00742892 {)}\\
        tmle        & 0.003217325   &{(} -0.002011248{{,}}  & 0.008445897{)}  & 0.002126624     &{(} -0.002878132{{,}}  & 0.007131379{)} \\ \hline
                    \multicolumn{7}{c}{The fifth year after treatment is considered for the outcome}           \\ \hline \hline
        naive       & -0.005438262  &\multicolumn{2}{c}{}  & -0.007293319 & \multicolumn{2}{c}{}  \\ 
        DRcnn       & 0.002465331   &{(} -0.003974954{{,}}  & 0.008905617{)}  & 0.001356161     &{(} -0.005572595{{,}}  & 0.008284916{)} \\
        DRmlp       & 0.009120043 &{(} 0.002080343{{,}} & 0.016159742{)} & 0.008610943 &{(} 0.001664695{{,}} & 0.01555719{)} \\ 
        ORds      & 0.001330909   &{(} -0.130924679{{,}}  & 0.133586497{)}  & 0.001974136     &{(} -0.098239946{{,}}  & 0.102188218{)} \\
        DRss     & 0.002021869   &{(} -0.003766136{{,}}  & 0.007809873{)}  & 0.002440018     &{(} -0.00326677 {{,}}  & 0.008146806{)} \\
        DRds      & 0.002706632   &{(} -0.003072861{{,}}  & 0.008486125{)}  & 0.002846952     &{(} -0.002859636{{,}}  & 0.00855354 {)}\\
        tmle        & 0.002038591   &{(} -0.003749006{{,}}  & 0.007826189{)}  & 0.002425458     &{(} -0.003281062{{,}}  & 0.008131978{)} \\ \hline
            \end{tabular}
\end{table}
\npnoround
\spacingset{1}

\begin{comment}

\bigskip
\begin{center}
{\large\bf SUPPLEMENTARY MATERIAL}
\end{center}

\begin{description}

\item[Title:] Brief description. (file type)

\item[R-package for  MYNEW routine:] R-package MYNEW containing code to perform the diagnostic methods described in the article. The package also contains all datasets used as examples in the article. (GNU zipped tar file)

\item[HIV data set:] Data set used in the illustration of MYNEW method in Section~ 3.2. (.txt file)

\end{description}

\section{BibTeX}

We hope you've chosen to use BibTeX!\ If you have, please feel free to use the package natbib with any bibliography style you're comfortable with. The .bst file Chicago was used here, and agsm.bst has been included here for your convenience. 
\end{comment}
\bibliographystyle{chicago}

\bibliography{cites.bib}

\begin{thebibliography}{}

\bibitem[\protect\citeauthoryear{Barban, de~Luna, Lundholm, Svensson, and
  Billari}{Barban et~al.}{2020}]{doi:10.1177/0049124117729697}
Barban, N., X.~de~Luna, E.~Lundholm, I.~Svensson, and F.~C. Billari (2020).
\newblock Causal effects of the timing of life-course events: Age at retirement
  and subsequent health.
\newblock {\em Sociological Methods \& Research\/}~{\em 49\/}(1), 216--249.

\bibitem[\protect\citeauthoryear{Bartlett, Bousquet, and Mendelson}{Bartlett
  et~al.}{2005}]{bartlett2005local}
Bartlett, P.~L., O.~Bousquet, and S.~Mendelson (2005).
\newblock Local rademacher complexities.
\newblock {\em The Annals of Statistics\/}~{\em 33\/}(4), 1497--1537.

\bibitem[\protect\citeauthoryear{Bartlett, Harvey, Liaw, and
  Mehrabian}{Bartlett et~al.}{2019}]{bartlett2019nearly}
Bartlett, P.~L., N.~Harvey, C.~Liaw, and A.~Mehrabian (2019).
\newblock Nearly-tight vc-dimension and pseudodimension bounds for piecewise
  linear neural networks.
\newblock {\em The Journal of Machine Learning Research\/}~{\em 20\/}(1),
  2285--2301.

\bibitem[\protect\citeauthoryear{Belloni, Chernozhukov, and Hansen}{Belloni
  et~al.}{2014}]{belloni2014inference}
Belloni, A., V.~Chernozhukov, and C.~Hansen (2014).
\newblock Inference on treatment effects after selection among high-dimensional
  controls.
\newblock {\em The Review of Economic Studies\/}~{\em 81\/}(2), 608--650.

\bibitem[\protect\citeauthoryear{Chernozhukov, Chetverikov, Demirer, Duflo,
  Hansen, Newey, and Robins}{Chernozhukov
  et~al.}{2018}]{chernozhukov2018double}
Chernozhukov, V., D.~Chetverikov, M.~Demirer, E.~Duflo, C.~Hansen, W.~Newey,
  and J.~Robins (2018).
\newblock Double/debiased machine learning for treatment and structural
  parameters.
\newblock {\em The Econometrics Journal\/}~{\em 21\/}(1), C1--C68.

\bibitem[\protect\citeauthoryear{Chernozhukov, Hansen, and
  Spindler}{Chernozhukov et~al.}{2016}]{hdm}
Chernozhukov, V., C.~Hansen, and M.~Spindler (2016).
\newblock {hdm}: High-dimensional metrics.
\newblock {\em R Journal\/}~{\em 8\/}(2), 185--199.

\bibitem[\protect\citeauthoryear{Farrell}{Farrell}{2015}]{Farrell_2015}
Farrell, M.~H. (2015, Nov).
\newblock Robust inference on average treatment effects with possibly more
  covariates than observations.
\newblock {\em Journal of Econometrics\/}~{\em 189\/}(1), 1–23.

\bibitem[\protect\citeauthoryear{Farrell}{Farrell}{2018}]{farrell2018robust}
Farrell, M.~H. (2018).
\newblock Robust inference on average treatment effects with possibly more
  covariates than observations.
\newblock {\em arXiv:1309.4686v3\/}.

\bibitem[\protect\citeauthoryear{Farrell, Liang, and Misra}{Farrell
  et~al.}{2021}]{farrell2018deep}
Farrell, M.~H., T.~Liang, and S.~Misra (2021).
\newblock Deep neural networks for estimation and inference.
\newblock {\em Econometrica\/}~{\em 89\/}(1), 181--213.

\bibitem[\protect\citeauthoryear{Kennedy}{Kennedy}{2016}]{kennedy2016semiparametric}
Kennedy, E.~H. (2016).
\newblock Semiparametric theory and empirical processes in causal inference.
\newblock In {\em He H., Wu P., Chen D.-G. (eds) Statistical causal inferences
  and their applications in public health research}, pp.\  141--167. Springer.

\bibitem[\protect\citeauthoryear{Klusowski and Barron}{Klusowski and
  Barron}{2018}]{klusowski2018approximation}
Klusowski, J.~M. and A.~R. Barron (2018).
\newblock Approximation by combinations of relu and squared relu ridge
  functions with $\ell^ 1$ and $\ell^ 0$ controls.
\newblock {\em IEEE Transactions on Information Theory\/}~{\em 64\/}(12),
  7649--7656.

\bibitem[\protect\citeauthoryear{LeCun, Bengio, et~al.}{LeCun
  et~al.}{1995}]{lecun1995convolutional}
LeCun, Y., Y.~Bengio, et~al. (1995).
\newblock Convolutional networks for images, speech, and time series.
\newblock {\em The handbook of brain theory and neural networks\/}~{\em
  3361\/}(10), 1995.

\bibitem[\protect\citeauthoryear{Lindgren, Nilsson, de~Luna, and
  Ivarsson}{Lindgren et~al.}{2016}]{simsam:2016}
Lindgren, U., K.~Nilsson, X.~de~Luna, and A.~Ivarsson (2016).
\newblock Data resource profile: Swedish microdata research from childhood into
  lifelong health and welfare ({U}me\aa\ {SIMSAM} {L}ab).
\newblock {\em International Journal of Epidemiology\/}~{\em 45}, 1075--1075.

\bibitem[\protect\citeauthoryear{Moosavi, H{\"a}ggstr{\"o}m, and
  de~Luna}{Moosavi et~al.}{2021}]{moosavi2021costs}
Moosavi, N., J.~H{\"a}ggstr{\"o}m, and X.~de~Luna (2021).
\newblock The costs and benefits of uniformly valid causal inference with
  high-dimensional nuisance parameters.
\newblock {\em To appear in Statistical Science. ArXiv preprint
  arXiv:2105.02071\/}.

\bibitem[\protect\citeauthoryear{Robins, Rotnitzky, and Zhao}{Robins
  et~al.}{1994}]{robins1994estimation}
Robins, J.~M., A.~Rotnitzky, and L.~P. Zhao (1994).
\newblock Estimation of regression coefficients when some regressors are not
  always observed.
\newblock {\em Journal of the American statistical Association\/}~{\em
  89\/}(427), 846--866.

\bibitem[\protect\citeauthoryear{Rubin}{Rubin}{1974}]{rubin1974estimating}
Rubin, D.~B. (1974).
\newblock Estimating causal effects of treatments in randomized and
  nonrandomized studies.
\newblock {\em Journal of educational Psychology\/}~{\em 66\/}(5), 688.

\bibitem[\protect\citeauthoryear{Scharfstein, Rotnitzky, and
  Robins}{Scharfstein et~al.}{1999}]{scharfstein1999rejoinder}
Scharfstein, D., A.~Rotnitzky, and J.~Robins (1999).
\newblock Rejoinder to comments on “adjusting for non-ignorable drop-out
  using semiparametric non-response models?”.
\newblock {\em Journal of the American Statistical Association\/}~{\em 94},
  1121--1146.

\bibitem[\protect\citeauthoryear{Stein}{Stein}{1970}]{10.2307/j.ctt1bpmb07}
Stein, E.~M. (1970).
\newblock {\em Singular Integrals and Differentiability Properties of Functions
  (PMS-30)}.
\newblock Princeton University Press.

\bibitem[\protect\citeauthoryear{Tan}{Tan}{2007}]{Tan:2007}
Tan, Z. (2007).
\newblock Comment: Understanding {OR}, {PS} and {DR}.
\newblock {\em Statistical Science\/}~{\em 22\/}(4), 560--568.

\bibitem[\protect\citeauthoryear{Tsiatis}{Tsiatis}{2006}]{tsiatis2006}
Tsiatis, A. (2006).
\newblock {\em Semiparametric theory and missing data}.
\newblock Springer Science \& Business Media.

\bibitem[\protect\citeauthoryear{van~der Laan and Rose}{van~der Laan and
  Rose}{2011}]{van2011targeted}
van~der Laan, M. and S.~Rose (2011).
\newblock {\em Targeted Learning: Causal Inference for Observational and
  Experimental Data}.
\newblock Springer Series in Statistics. Springer New York.

\bibitem[\protect\citeauthoryear{Zhou}{Zhou}{2020}]{zhou2020universality}
Zhou, D.-X. (2020).
\newblock Universality of deep convolutional neural networks.
\newblock {\em Applied and computational harmonic analysis\/}~{\em 48\/}(2),
  787--794.

\end{thebibliography}
\end{document}